\documentclass[10pt,journal,compsoc,final]{IEEEtran}
\pdfoutput=1
\usepackage{graphicx}
\usepackage{booktabs}
\usepackage{color}
\usepackage{array}
\usepackage{float}
\usepackage{hyperref}
\hypersetup{colorlinks=true, urlcolor=blue, linkcolor=red, citecolor=green}
\usepackage{multirow}
\usepackage{amsmath}
\usepackage{amssymb}
\usepackage{colortbl} 
\usepackage{bibspacing}

\usepackage{review}

\setcoverletter{CoverLetterR1.tex}
\setrevision{1}

\definecolor{visioncolor}{rgb}{0.57,0.815,0.314} 
\definecolor{audiocolor}{rgb}{1.,1.,0.} 
\definecolor{ircolor}{rgb}{1.,0.,0.} 
\definecolor{btcolor}{rgb}{0.,0.688,0.938} 
\definecolor{accelcolor}{rgb}{0.55,0.348,0.699}

\def\eg{\textit{e.g.}}
\def\ie{\textit{i.e.}}

\def\etal{\textit{et al.}}
\def\etc{\textit{etc.}}

\newlength{\cheating}
\begin{document}
\title{SALSA: A Novel Dataset for Multimodal Group Behavior Analysis}
\author{Xavier~Alameda-Pineda, Jacopo~Staiano, Ramanathan~Subramanian,~\IEEEmembership{Member,~IEEE,} Ligia~Batrinca, Elisa~Ricci,~\IEEEmembership{Member,~IEEE,} Bruno~Lepri, Oswald~Lanz,~\IEEEmembership{Member,~IEEE,} Nicu~Sebe,~\IEEEmembership{Senior Member,~IEEE}%
\IEEEcompsocitemizethanks{\IEEEcompsocthanksitem Xavier Alameda-Pineda, Ligia Batrinca and Nicu Sebe are with the Department of Information 
Engineering and Computer Science, University of Trento, Trento, Italy. (Email: xavier.alamedapineda,  ligia.batrinca, 
niculae.sebe@unitn.it).\protect%
\IEEEcompsocthanksitem Elisa~Ricci, Bruno Lepri and Oswald Lanz are with Fondazione Bruno Kessler, Trento,  Italy. (Email: eliricci, lepri, lanz@fbk.eu) \protect%
\IEEEcompsocthanksitem Jacopo-Staiano is with Sorbonne Universit\'ees, UPMC, CNRS, LIP6, France (Email: jacopo.staiano@lip6.fr) \protect%
\IEEEcompsocthanksitem Ramanathan Subramanian is with the Advanced Digital Sciences Center,  University of Illinois at Urbana-Champaign, Singapore. 
(Email: subramanian.r@adsc.com.sg)}\vspace{-0.7cm}}

\maketitle
$ $\vspace{-0.8cm}\\
\begin{abstract}
Studying free-standing conversational groups (FCGs) in unstructured social settings (\eg, \textit{cocktail party}) is gratifying due to the wealth of 
information available at the \textit{\textbf{group}} (mining social networks) and \textit{\textbf{individual}} (recognizing native behavioral and 
personality traits) levels. However, analyzing social scenes involving FCGs is also highly \addnote[chal1]{1}{challenging due to the difficulty in extracting behavioral cues such as target locations, their speaking activity and head/body pose due to crowdedness and presence of extreme occlusions}. To this end, we propose \textbf{SALSA}, a novel dataset facilitating multimodal and Synergetic sociAL Scene Analysis, and make 
two main 
contributions to research on automated social interaction analysis: (1) SALSA records social interactions among 18 participants in a natural, indoor 
environment for over 60 minutes, under the \textit{poster presentation} and \textit{cocktail party} contexts presenting difficulties in the form of 
low-resolution images, lighting variations, numerous occlusions, reverberations and interfering sound sources; (2) To alleviate these problems we 
facilitate multimodal analysis by recording the social interplay using four \textit{static surveillance cameras} and \textit{sociometric badges} worn 
by each participant, comprising the \textit{microphone}, \textit{accelerometer}, \textit{bluetooth} and \textit{infrared} sensors. In addition to raw 
data, we also provide \textit{annotations} concerning individuals' personality as well as their \textit{position}, \textit{head, body orientation} 
and \textit{F-formation} information over the entire event duration. Through extensive experiments with state-of-the-art approaches, we show (a) the limitations 
of current methods and (b) how the recorded multiple cues synergetically aid automatic analysis of social interactions. SALSA is available at 
\texttt{http://tev.fbk.eu/salsa}.
\end{abstract}
\begin{keywords}
Multimodal group behavior analysis, Free-standing conversational groups, Multimodal social data sets, Tracking, Speaker recognition, Head and body pose, F-formations, Personality traits.
\end{keywords} 

\vspace{\cheating}

\section{Introduction}
\IEEEPARstart{H}{umans} are social animals by nature, and the importance of \textbf{social interactions} for our physical and mental well-being has 
been widely acknowledged. Therefore, it is unsurprising that social interactions have been studied extensively by social psychologists in a variety of 
contexts. Fundamental research on social interactions was pioneered by Goffman~\cite{goffman1961}, whose \textit{symbolic interaction perspective} 
explains society via the everyday behavior of people and their interactions. Face-to-face conversations are the most common form of social 
interactions, and \textbf{free-standing conversational groups} (FCGs) \addnote[stress-f-for]{1}{denote small groups of two or more co-existing persons 
engaged in ad-hoc interactions~\cite{goffman1966behavior}.  FCGs emerge naturally in diverse social occasions, and interacting persons are 
characterized by mutual scene locations and poses resulting in spatial patterns known as F-formations~\cite{kendon1990}\footnote{A F-formation is a 
set of possible configurations in space that people may assume while participating in a social interaction.}}. Also, social behavioral cues like how 
much individuals speak and attend to others are known to be correlated with individual and group-specific traits such as Extraversion~\cite{Ashton02} 
and Dominance~\cite{Rosa79}. 

\begin{figure}[t]
\includegraphics[width=0.4\linewidth,height=3cm]{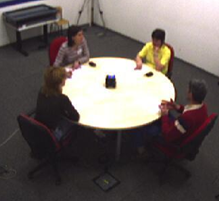}\hspace{0.02in}%
\includegraphics[width=0.57\linewidth,height=3cm]{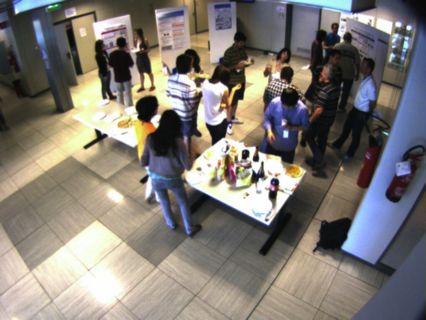}
\caption{\label{ill} Analysis of round-table meetings (left, adopted from \textit{Mission Survival} dataset~\cite{zancanaro2006automatic}) has been attempted extensively-- orderly spatial arrangement 
and sufficient separation between persons enable reliable extraction of behavioral cues for each person from such scenes.  Exemplar SALSA frame (right) showing FCGs-- varying illumination, 
low resolution of faces, extreme occlusions and crowdedness makes AASI highly challenging (best-viewed in color and under zoom).}
\vspace{-0.6cm}
\end{figure}

Automated analysis of social interactions (AASI) has been attempted for over a decade now. The ability to recognize social interactions and infer 
social cues is critical for a number of applications such as surveillance, robotics and social signal processing. Studying unstructured social scenes 
(\eg, a cocktail party) is extremely challenging since it involves inferring \addnote[chal2]{1}{(i) locations and head/body 
orientations of targets (persons) in the scene, (ii) semantic and prosodic auditory content corresponding to each target, (iii) F-formations and persons interacting at a particular time (addresser and 
addressee recognition), (iv) attributes such as the big-five personality traits~\cite{John1999} from the above social behavioral cues}.

Some of the aforementioned problems have been effectively addressed in controlled environments such as round-table meetings (Fig.~\ref{ill} (left)), where behavioral cues 
concerning orderly arranged participants can be reliably acquired through the use of web-cameras and close-talk microphones. However, \textit{all} of 
them remain unsolved in unstructured social settings, where only distant surveillance cameras can be used to capture FCGs~\cite{kendon1990}, and 
microphones may be insufficient to clearly recognize the speaker(s) at a given time instant due to 
scene crowding. We are expressly interested in analyzing FCGs (Fig.~\ref{ill} (right)) in this work.
 
To address the challenges involved in analyzing FCGs, we present \textbf{SALSA}, a novel \addnote[public]{1}{dataset\footnote{The SALSA dataset (raw 
data, annotations and associated code) is available at \texttt{http://tev.fbk.eu/salsa}.}}~facilitating multimodal and Synergetic 
sociAL Scene Analysis. In contrast to social datasets studying round-table meetings~\cite{zancanaro2006automatic,mccowan2005ami}, or examining FCGs on 
a small scale~\cite{setti2014,zen2010space}, SALSA contains uninterrupted recordings of an indoor social event involving 18 subjects over 60 minutes, 
thereby serving as a rich and extensive repository for the behavioral analysis and social signal processing communities. In addition to the raw 
multimodal data, SALSA also contains position, pose and F-formation annotations over the entire event duration for evaluation purposes, as well as 
information regarding participants' personality traits.

The social event captured in SALSA comprised two parts: (1) a \textit{poster presentation} session, and (2) a \textit{cocktail party} scene where participants freely interacted with each other. 
There were no constraints imposed in terms of how the subjects were to behave (\ie, no scripting) or the spatial layout of the scene during the recordings. Furthermore, the indoor environment where 
the event was recorded was prone to varying illumination and reverberation conditions. \addnote[constrained-f-for]{1}{The geometry of F-formations is 
also influenced by the physical  space where the social interaction is taking place, and the poster session was intended to simulate a 
semi-structured social setting facilitating speaker labeling as described in Section~\ref{sec:exp:speaker}. While it is reasonable to expect observers 
to stand in a semi-circular fashion around the poster presenter, none of the participants were instructed on where to stand or what to attend}. 
Finally, as seen in Fig.~\ref{ill} (right), the crowded nature of the scene and resulting F-formations 
gave rise to extreme occlusions, which along with the low-resolution of faces captured by surveillance cameras, make visual analysis extremely 
challenging. Also, crowdedness poses difficulties to audio-based analysis for solving problems such as speaker recognition. Overall, SALSA represents 
the most challenging dataset for studying FCGs to our knowledge.

To alleviate the difficulties in traditional audio-visual analysis due to the challenging nature of the scene, in addition to four wall-mounted cameras acquiring the scene, \textit{sociometric badges}~\cite{choudhury2003} 
were also worn by participants to record various aspects of their behavior. These badges include a microphone, an accelerometer, bluetooth and infrared (IR) sensors. The microphone records auditory  
content that can be used to perform speaker recognition under noisy conditions, while the accelerometer captures person motion. Bluetooth and IR transmitters 
and receivers provide information regarding interacting persons, and are useful for inferring body pose under occlusions. Cumulatively, these sensors can synergetically enhance estimation of target locations and their head and body orientations, face-to-face interactions, F-formations, and thereby provide a rich description of FCGs' behavior. Through the SALSA dataset, we make two main contributions to AASI research:

\begin{itemize}
 \item Firstly, we believe that the challenging nature of SALSA will enable researchers to appreciate the limitations of current AASI approaches, and spur focused and intensive 
research in this regard. 
 \item We go beyond audio-visual analysis for studying FCGs, and show how multimodal analysis can alleviate difficulties posed by occlusions and crowdedness to more precisely estimate various 
behavioral cues and personality traits therefrom.
\end{itemize}

The paper is organized as follows. Section~\ref{LR} reviews prior unimodal and multimodal approaches to AASI, while Section~\ref{RG} highlights the limitations of traditional AASI 
approaches and datasets, motivating the need for SALSA. Section~\ref{DS} describes the SALSA dataset, and the synergistic benefit of employing audio, visual and 
sensor-based cues is demonstrated via experiments in Section~\ref{Exp}. We finally conclude in Section~\ref{Con}.

\vspace{\cheating}
\section{Literature Review}\label{LR}

This section reviews the state-of-the-art in social behavior analysis with specific emphasis on methods analyzing FCGs. We will first discuss unimodal approaches (\ie,  
vision-, audio- and wearable sensor-based) and then describe expressly multimodal approaches.

\vspace{\cheating}
\subsection{Vision-based approaches}
Challenges pertaining to surveillance scenes involving FCGs addressed by vision-based methods include detection and tracking of targets in the scene, estimation of social attention direction and detection of F-formations. We describe works that have examined each of the above problems as follows.

Given the cluttered nature of social scenes involving FCGs, detecting and tracking the locations of multiple targets is in itself a highly challenging 
task. Multi-target tracking from monocular images is achieved through the use of a dynamic Bayesian network in~\cite{smith2008tracking}. As extreme 
occlusions are common in social scenes involving FCGs (see Fig.~\ref{ill}), employing information from multi-camera views can enable robust tracking. 
The color-based particle filter tracker proposed in~\cite{lanz2006approximate} for multi-target tracking with explicit occlusion handling is 
notable in this regard. Of late, tracking-by-detection~\cite{chen2012we} combined with data association using global appearance and motion 
optimization~\cite{Berclaz11} has benefited multi-target tracking. Some works have further focused on identity-preserving tracking over long 
sequences~\cite{shitrit2014multi}. These methods assume a sufficiently large number of high-confidence detections which might not always be available 
with FCGs due to persistent long-term occlusions, although recent works have focused on the detection problem under partial occlusion 
\cite{mathias2013handling,wojek2011cvpr}. Multi-target tracking is further shown to improve by incorporating aspects of human social 
behavior~\cite{eth_biwi_00645,ge2012pami}, as well as with activity analysis~\cite{lan2010nips}. However, F-formations are special groups, 
 characterized by static arrangements of interacting persons constrained by their locations and head/body orientations.


\indent Social attention determination in round-table meetings, where high-resolution face images are available, has been studied extensively. Vision-based approaches typically employ head pose as a 
proxy to determine social attention direction~\cite{stiefelhagen2002tracking}. In comparison, head pose estimation (HPE) from blurry surveillance videos is much 
more difficult. Visual attention direction of moving targets is estimated by Smith \etal~\cite{smith2008tracking} using position and head pose cues, but social scenes or occlusions are not addressed 
here. Some works have exploited the relationship between walking direction and head/body pose to achieve HPE from surveillance videos under limiting conditions where no labeled training 
examples are available, or under occlusions~\cite{benfold,chen2012we}. Focus-of-attention estimation in social scenes is explicitly addressed by 
Bazzani \etal~\cite{bazzani2013social}, who model a person's visual field in a 3D scene using a subjective view frustum. 

\addnote[choi]{1}{Recently, the computer vision community has showed some interest in the detection and analysis of dyadic 
interactions~\cite{Jimenez,Patron-Perez2012} and more general groups~\cite{choi14discovering,eichner2010we,yang2012recognizing}}.
Also, the interest on detecting social interactions and F-formations from video has intensively grown. 
 Cristani~\etal~\cite{cristani2011,SettiHC13,setti2013multi} employ positional and head orientation estimates to detect F-formations
based on a Hough voting strategy. Bazzani \etal~\cite{bazzani2013social} gather information concerning  F-formations in a social scene using the Inter-relation pattern matrix. F-formations are modeled 
as maximal cliques in edge-weighted graphs \addnote[hung]{1}{in~\cite{hung2011detecting}}, and each target is presumed to be oriented towards the 
closest neighbor but head/body orientation is not explicitly computed. 
Gan \etal~\cite{gan2013} detect F-formations upon inferring location and orientation of subjects using depth sensors and cameras. Setti \etal~\cite{setti2014} propose a graph-cuts 
based framework for F-formation detection based on the position and head orientation estimates of targets. Based on experiments performed on four 
datasets (\textit{IDIAP Poster}~\cite{hung2011detecting}, \textit{Cocktail Party}~\cite{zen2010space}, \textit{Coffee Break}~\cite{cristani2011} and 
\textit{GDet}~\cite{bazzani2013social}), their method outperforms six state-of-the-art methods in the literature.

\vspace{\cheating}

\subsection{Audio-based approaches}
Studying interactional behavior in unstructured social settings solely using audio or speech-based cues is extremely challenging, as FCGs are not only characterized by speaking activity, but also by non-verbal cues such as head and body 
pose, gestural and proxemic cues. Furthermore, classical problems in audio analysis become extremely challenging and remain unexplored in crowded indoor environments involving a large 
number of targets. Indeed, current methodologies for speaker diarization \cite{anguera2012}, sound source separation \cite{badeau2013} or localization \cite{alameda2014} address scenarios with few persons. Nevertheless, a few studies on audio-based detection of FCGs have been published.
Wyatt \etal~\cite{wyatt2011} tackle the problem of detecting FCGs upon recognizing the speaker using temporal and spectral audio features. Targets are then clustered to determine co-located groups on the basis of speaker labels. More recently, FCG detection and network inference is achieved in~\cite{luo2013} employing a conversation sensing system to perform speaker recognition, and F-formations are detected based on proximity information obtained using bluetooth sensors.

\vspace{\cheating}
\subsection{Wearable-sensor based approaches}\label{wearable}
Wearable sensors can provide complementary behavioral cues in situations where visual and speech data are unreliable due to occlusions and crowdedness. Hung \etal~\cite{hung2014} detect FCGs in social gatherings by measuring motion via an accelerometer. With the increased usage of smartphones, mobile sensor data 
have also become a viable choice for analysis of social interactions or more complex social systems~\cite{eagle2006}. Via mobile phones, proximity can be inferred from Wifi and 
bluetooth~\cite{madan2012}. However, the spatial resolution of these sensors is limited to only a few meters, and the co-location of mobile devices does not necessarily indicate a social interaction 
between the corresponding individuals. Therefore, Cattuto \etal~\cite{cattuto} propose a framework that balances scalability and resolution through a sensing tier consisting of cheap and unobtrusive 
active RFID devices, which are capable of sensing face-to-face interactions as well as spatial proximity over different scale lengths down to one meter or less. Nevertheless, we note that many of these works address relatively less crowded scenarios, not comparable in complexity to SALSA.

\vspace{\cheating}
\subsection{Multimodal approaches}
Multimodal approaches to social interaction have essentially examined small-group interactions such as round-table meetings, and mainly involve audio-visual analysis as detailed below. Examples of 
databases containing audio-visual recordings and associated annotations are the Canal9~\cite{vinciarelli2009canal9}, AMI~\cite{mccowan2005ami}, Mission Survival~\cite{zancanaro2006automatic}, Free 
Talk \cite{kurtic2012} 
and the Idiap WOLF~\cite{hung2010idiap} corpora. \addnote[addedcitations]{1}{In addition, the IMADE~\cite{sumi2010analysis} and UEM~\cite{mase2006ubiquitous} technological frameworks for recording of multimodal data to describe social scenes involving FCGs are available}.
All these data collection efforts have inspired inter-disciplinary research in the field of human behavior understanding and led to the emergence of the social signal processing community~\cite{vinciarelli2009social}. Several of these databases have been utilized for isolating traits 
relating to an individual (\eg, big-five personality traits~\cite{subramanian2013relationship}) or a group (such as \emph{dominance}~\cite{hung2007using}).

Choudhury and Pentland~\cite{choudhury2003} initiated behavior analysis using wearable sensors by developing the \textit{Sociometer}. Recently, Olgu\'in \etal~\cite{olguin2009} proposed the \textit{Sociometric badge}, which stores (i) motion using an accelerometer, (ii) speech features (rather than raw audio),  (iii) position information and (iv) proximity to other individuals 
using a bluetooth sensor, and (v) face-to-face interactions via an infrared sensor. Sociometric badges have been used to capture face-to-face communication patterns, examine relationships among individuals and model collective and organizational behavior~\cite{lepri2012socio}, detect personality traits and 
states~\cite{subramanian2013relationship}, and predict outcomes such as productivity and job satisfaction~\cite{olguin2010}. Another notable AASI work
employing multi-sensory information is that of Matic \etal~\cite{matic2012}, who estimate body orientation 
and inter-personal distance via mobile data and speech activity to detect social interactions.

\begin{figure}[t]
    \centering
     \includegraphics[width=.9\linewidth]{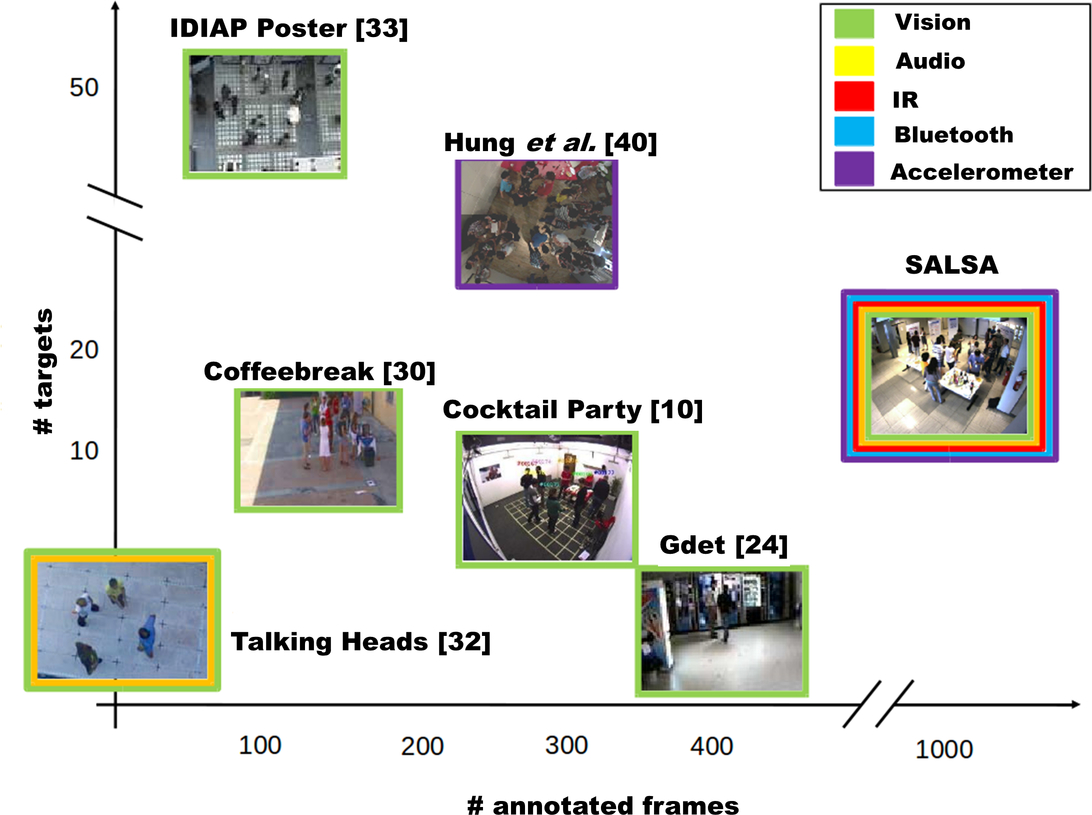}\vspace{-.1in}
     \caption{
		Existing datasets facilitating F-formation detection. 
Frame border colors encode sensing modalities.}
    \label{fig:FFPic}
    \vspace{-0.4cm}
\end{figure}

\vspace{\cheating}
\section{Spotting the research gap}\label{RG}
While human behavior has been studied extensively in controlled settings such as round-table meetings, achieving the same with FCGs is way more 
difficult as close audio-visual examination of targets is precluded by the crowded and occluded nature of the scene. We carried out an extensive analysis of previous AASI 
data sets focusing on FCGs; they have mainly been used to address two research problems: (1) Detecting F-formations and (2) Studying individual and 
group behavior from multiple sensing modalities. 

The vast majority of works addressing F-formation detection are vision-based. Fig.~\ref{fig:FFPic} presents snapshots of datasets used for F-formation detection, and positions them with respect to the number (denoted using \#)
of annotated frames and scene targets. Among them, SALSA is unique due to its (i) multimodal nature, (ii) extensive annotations available over a long duration and (iii) challenging nature of the captured scene.

\begin{figure}[t]
    \centering
    \includegraphics[width=.9\linewidth]{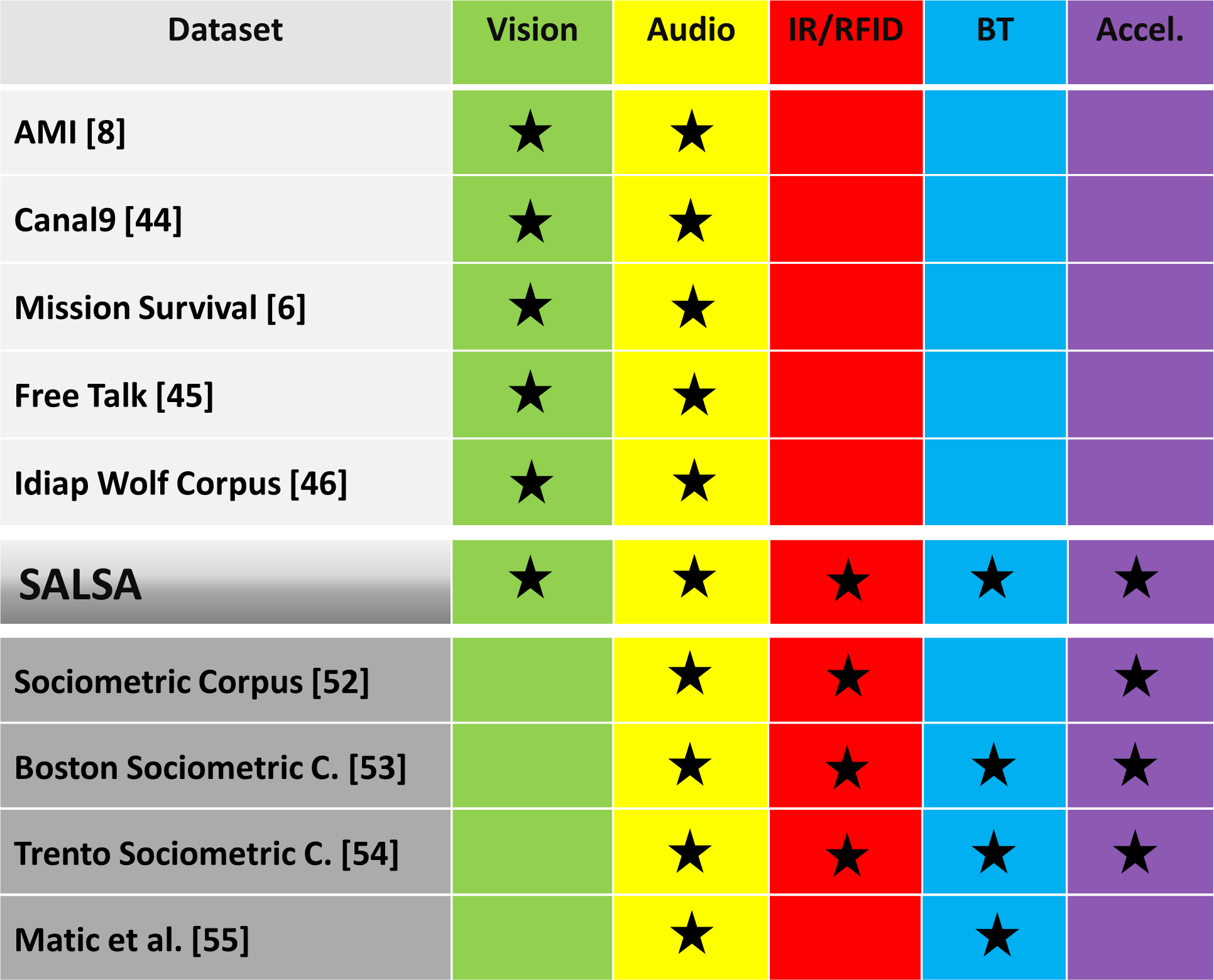}
    \caption{Datasets for social interaction analysis: the first five consist on round-table meetings and span over hours, while the last four study social networks/behavior and span over days/months.}
    \label{fig:MMPic}
    \vspace{-0.4cm}
\end{figure}

Figure~\ref{fig:MMPic} depicts the datasets used for individual and group behavioral analysis. While the first group (light-gray) consists of audio-visual recordings spanning over hours acquired under controlled settings, the second group (dark-gray) comprises datasets acquired over days/months for studying social networks and group relationships. 
SALSA again stands out as it records information from both static cameras and wearable sensors, leading to a previously non-existent and highly informative combination of sensing modalities. This section details some of the limitations of current AASI approaches regarding F-formation detection and individual and group behavior analysis, thereby throwing light on how SALSA can spur critical research in these respects.

\vspace{\cheating}
\subsection{Human tracking and pose estimation} 
Human tracking and pose estimation in a social context is challenging for several reasons. Firstly, a person's visual appearance can change considerably 
across the scene due to camera perspective and uneven illumination, as well as with pose and posture. Secondly, large and persistent 
occlusions are frequent in such scenes, which corrupt subsequent observations. Thirdly, integrating auditory information to aid localization and 
orientation estimation is also difficult due to its intermittent nature and the adverse impact of reverberations and interfering sources. Beyond 
inherent complexity, the state-of-the-art is further challenged when raw audio-visual data cannot be recorded for processing or scene-specific optimization, \eg, due to privacy concerns. 

Accurate FCG behavior analysis requires the correct assignment of observations to sources (targets) over the long run. Identity switches during tracking can corrupt the extraction of aggregated features that develop over time to infer personality traits, functional roles, or 
interaction networks, and long-term identity-preserving multi-target tracking is still unachievable. Furthermore, existing appearance-based pose estimation methods are not adapted to highly cluttered scenes with large and persistent occlusions. 
Finally, from the computational perspective, multi-target approaches are not able to robustly and efficiently scale to large groups.


\vspace{\cheating}
\subsection{Speech processing}
While numerous research studies have attempted speech, speaker and prosodics recognition under controlled conditions, several issues arise when auditory information is captured via mobile microphones 
in crowded indoor scenes. Firstly, regular indoor environments are prone to reverberations, which adversely affect many sound processing techniques. Secondly, intermittence of the speech signal 
necessitates speaker diarization prior to processing. 
Thirdly, the speech signal is also spatially sparse, and source separation techniques are usually required to segment speaker activity. Currently, there are no algorithms addressing source separation or diarization in the presence of a large number of sound sources and uncontrolled conditions. While multimodal approaches have addressed these problems via audio-visual processing, they still cannot work with large groups of people and crowded indoor environments involving unconstrained and evolving interactions.

\vspace{\cheating}
\subsection{F-formation detection}
Detecting F-formations in unconstrained environments is a complex task. As F-formations are characterized by mutually located and oriented persons, robust tracking and pose estimation algorithms are 
necessary. However, both multi-target tracking and head/body pose estimation in crowded scenes are difficult as discussed earlier. Even under ideal conditions, F-formation shapes are influenced by (i) 
the environment's layout, \ie, room shape, furniture and other physical obstacles, (ii) scene crowdedness and (iii) attention hotspots such a poster, painting, \etc~While existing methodologies 
typically assume that F-formation members are placed on an ellipse, robust F-formation detection requires accounting for the above factors as well. Also, most algorithms are visually driven, and few multimodal approaches exist to this end.

\vspace{\cheating}
\subsection{Inferring personality traits}
Works seeking to recognize personality traits from interactive behavior have traditionally relied on the visual and auditory modalities. Target position and pose, prosodic and intonation inference, face-to-face 
interaction detection, bodily gestures and facial expressions are commonly used for assessing the big-five personality traits. Most prior works study these behaviors in the context of round-table 
meetings, where participants are regularly arranged in space, and therefore their positions and body orientations are known \textsl{a priori}. Under these conditions, behavior analysis algorithms 
deliver precise estimates concerning interactional behavior, thereby facilitating personality trait recognition. Nevertheless, there are very few works that studied personality inference from 
unstructured interactions involving FCGs.

Assessing the personality traits of a large number of interacting persons in crowded scenarios, where the group structure evolves progressively, is a highly challenging task. In such cases, people constantly leave and join groups, and therefore groups are created, split, and merged. An in-depth analysis should take the group dynamics into account in addition to the evolving physical arrangements and occasional conversations. Given that personality inference is a sophisticated and subtle task, the right combination of cues extracted from different modalities can lead to a robust assessment. 

\vspace{\cheating}
\subsection{The \textit{raison d'\^etre} of SALSA}
Upon analyzing the state-of-the-art in behavior analysis and personality inference, we conclude that: (i) Even if multimodal analysis has been found 
to outperform unimodal approaches and provide a richer representation of social interplays, some key tasks are not yet addressed in a multimodal 
fashion, \eg pose estimation and F-formation detection; (ii) Social interactions have been studied under controlled settings, and there is a paucity 
of methods able to cope with unconstrained  environments involving large groups, crowded spaces and highly dynamic interactions, and (iii) Most 
existing approaches have independently studied the different behavioral tasks -- while it is known that bidirectional links between the tasks exist, 
these links have been rarely exploited. \addnote[taskssalsa]{1}{For instance, knowing the head and body orientations of individuals can help in the 
estimation of F-formations and vice-versa. Similarly, F-formation detection clearly benefits from accurate tracking algorithms, which at their turn 
can be influenced by the robust detection of F-formations}. \addnote[datadriven]{1}{In order to foster the study of the aforementioned challenges}, 
we recorded SALSA, whose description is presented in the next section.

\section{The SALSA data set} \label{DS}
In order to provide a new and challenging evaluation framework for novel methodologies addressing the aforementioned challenges, we introduce the SALSA (Synergetic sociAL Scene Analysis) dataset. 
SALSA represents an excellent test-bed for multimodal human behavior understanding due to the following reasons. Firstly, all behavioral data were collected in a regular indoor space with the 
participants only requiring to wear portable and compact sociometric badges which ensured naturalistic social behavior. Secondly, due to the unconstrained nature of the scene, the recordings contain numerous artifacts such as varying illumination, visual occlusions, reverberations or interfering sound sources. Thirdly, the recorded event involved 18 persons: such large social groups have rarely been studied in the behavior analysis literature. These participants did not receive any special instructions or scripts prior to the recording, and the resulting social interactions were therefore free-willed and hedonistic in nature. Finally, the social interplay was recorded via four wall-mounted surveillance cameras and the \textit{Sociometric 
badges}\footnote{\texttt{http://www.sociometricsolutions.com/}} worn by the targets. These badges recorded different aspects of the targets' social behavior such as audio or motion as detailed later. This combination of static cameras and wearable sensors is scarce in the literature, and provides a wealth of behavioral information as shown in Section~\ref{Exp}. These four salient characteristics place SALSA in a unique position among the various 
datasets available for studying social behavior, see Figures~\ref{fig:FFPic} and~\ref{fig:MMPic}.

\subsection{Scenario and roles}
\label{sec:scenario}
SALSA was recorded in a regular indoor space and the captured social event involved 18 participants and consisted of two parts of roughly equal duration. The first part consisted of a \textit{poster presentation} session, where four research studies were presented by graduate students. A fifth person chaired the poster session. In the second half, all participants were allowed to freely interact over food and beverages during a \textit{cocktail party}. 

It needs to be noted here that while some participants had specific roles to play during the poster presentation session, none were given any instructions on how to act in the form of a script. Consequently, the interaction dynamics correspond to those of a natural social interplay. Obviously, participants with different roles (chair, poster presenter, attendee) are expected to have different interaction dynamics, and these roles were designed to help behavioral researchers working on role recognition.

\setlength\extrarowheight{2pt}
\begin{table}[t]
 \centering
 \caption{\label{tab:sensors}Description of the sensors used in SALSA. STFT denotes short-time Fourier transform.}\vspace{-.15in}
 \begin{tabular}{lll}
    Sensor & Output & Freq. (Hz) \\
    \rowcolor{visioncolor} Vision & 4 synchronized images & $15$ \\
    \rowcolor{audiocolor} Audio & Amplitude stats \& STFT & $2$ \& $30$ \\
    \rowcolor{ircolor} Infra-red & Detected badge's ID & $1$ \\
    \rowcolor{btcolor} Bluetooth & Detected badge's ID & $1/60$\\
    \rowcolor{accelcolor} Accelerometer & Body motion & $20$\\
 \end{tabular}
 \vspace{-0.4cm}
\end{table}
\setlength\extrarowheight{0pt}

\subsection{Sensors}
\label{sec:sensors}
The SALSA data were captured by a camera network and wearable badges worn by targets. The camera network comprised four synchronized static 
RGB cameras (\addnote[resolution]{1}{$1024\times768$}~resolution) operating at 15 frames per second (fps). Each participant wore a sociometric badge 
during the 
recordings which is a $9\times 6 \times 0.5$~cm box equipped with four sensors, namely, a microphone, an infrared (IR) beam and detector, a 
Bluetooth detector and an accelerometer. The badges are battery-powered and store recorded data on a USB card without the need for any wired 
connection, thus enabling natural social interplay. Table~\ref{tab:sensors} presents an overview of the five sensors used.

\vspace{\cheating}
\subsection{Ground truth data} \label{GTdata}
\subsubsection*{Annotations}  \label{Annot}
In order to fulfill the requirements expected of a systematic evaluation framework, SALSA provides ground-truth annotations, which were performed 
either manually or semi-automatically over the entire event duration. The annotations were produced in two steps. In the first step, using a dedicated multi-view scene annotation tool, the {\it position}, {\it head} and {\it body} orientation of each target was annotated every 45 frames (3 seconds). 
\addnote[annotators]{1}{To speed up the annotation process, the total number of targets was divided among three annotators. A target's position, head 
and body orientation were annotated by a first annotator and then double-checked by the second. Discrepancies between their judgments were resolved by 
a third annotator. All annotators were clearly instructed on how to perform the annotations}. To facilitate the annotation task, markings from the 
previous annotated frame were displayed so that only small modifications were needed.

\begin{figure}[t]
    \centering
    \includegraphics[width=1.8in]{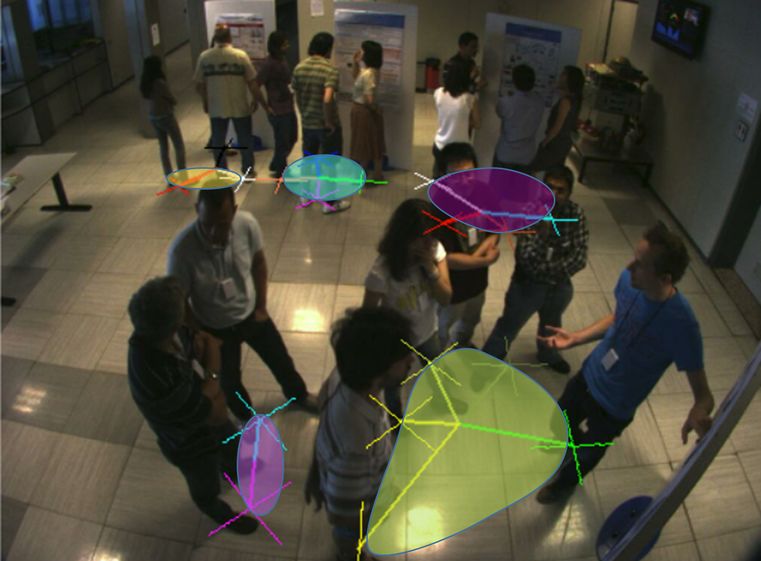}
    \caption{Five annotated F-formations represented via connections between feet positions (crosses) of interacting targets. Corresponding O-spaces are denoted by the colored convex shapes. }
    \label{fig:annot1}
    \vspace{-0.4cm}
\end{figure}

In the second step, annotated positions and head/body orientations were used for deducing F-formations. Annotations were again performed every 45 frames and we employed the following criteria for detecting F-formations: \addnote[describe-f-for]{1}{an F-formation is characterized by the mutual locations and head, body orientations of interacting targets, and is defined by the convex \textit{\textbf{O-space}} they encompass such that each target has unhindered access to its center. A valid F-formation was assumed if the constituent targets were in one of the established patterns, or had direct and unconstrained access to the O-space center in case of large groups (refer to~\cite{cristani2011} for details)}. Figure~\ref{fig:annot1} illustrates five annotated F-formations around four posters \addnote[annotationfigure]{1}{(target feet positions are marked with crosses) and corresponding O-spaces. Considering the two groups in the foreground, the F-formation in front of the poster on the right does not include the FCG with two targets on the left, since neither of them have access to the center of the larger group.}

\vspace{\cheating}
\subsubsection*{Personality data}
\begin{figure}[t]
\centering
 \includegraphics[width=0.9\columnwidth]{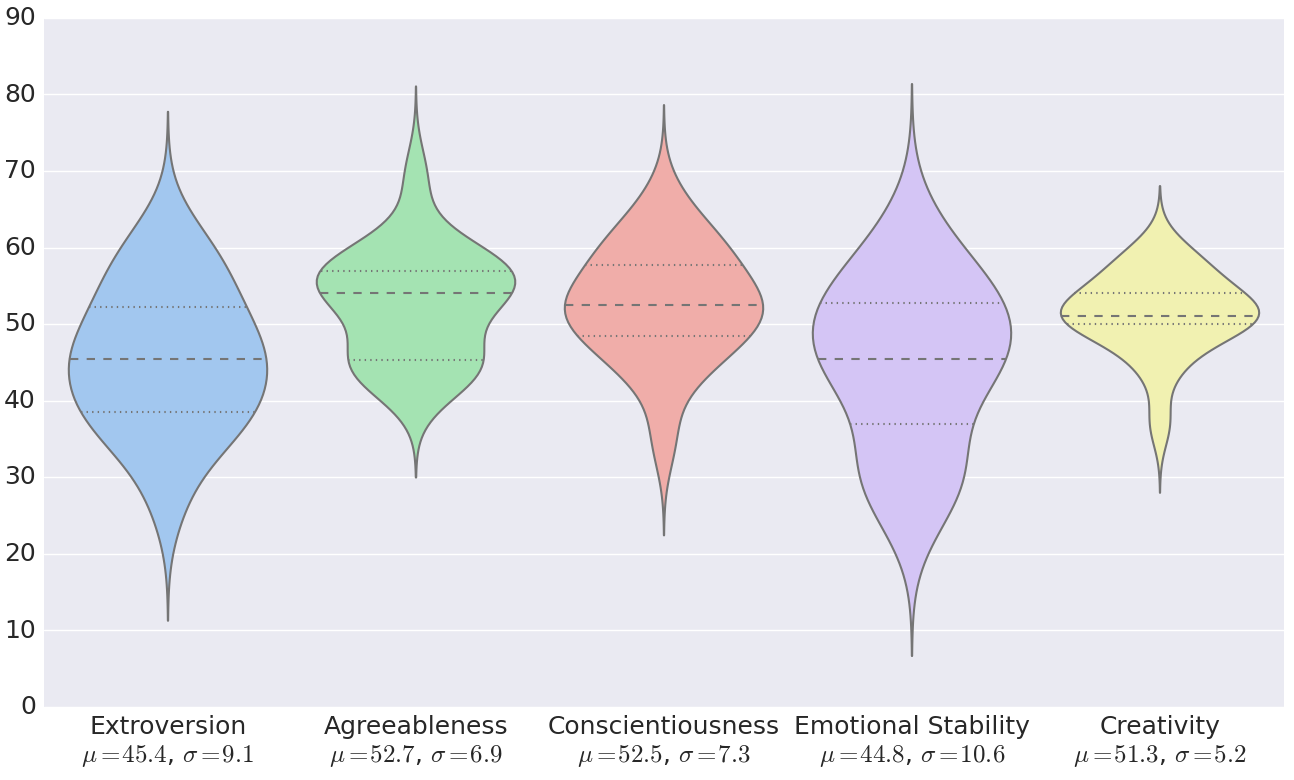}
 \caption{Distributions of the big-five personality traits.}\vspace{-.1in}
 \label{fig:pers}
 \vspace{-0.4cm}
\end{figure}

SALSA also contains big-five personality trait scores of participants to facilitate behavioral studies. Prior to data collection, all participants filled the \textit{Big Five} personality questionnaire~\cite{John1999}. The Big Five 
questionnaire owes its name to the five traits it assumes as constitutive of personality: \textit{Extraversion}-- being sociable, assertive, playful vs. aloud, reserved, shy; \textit{Agreeableness}-- being friendly and cooperative vs. antagonistic and fault-finding; \textit{Conscientiousness}-- being self-disciplined, organized vs. inefficient, careless; \textit{Emotional Stability}-- being calm and equanimous vs. insecure and anxious; and \textit{Creativity}-- being intellectual, insightful vs. shallow, unimaginative. In the questionnaire, each trait is investigated via ten items assessed on a 1--7 Likert scale. The final trait scores were computed according to the 
procedure detailed in~\cite{perugini2002}, and the distributions of these traits over the 18 targets are presented in Figure \ref{fig:pers}.

\vspace{\cheating}
\section{Experiments on SALSA} \label{Exp}
This section is devoted to evaluate the performance of state-of-the-art methodologies for different behavioral tasks on SALSA. In order to ensure reproducibility of results, we will provide (i) the dataset, (ii) the software used to produce the obtained results and (iii) a detailed description of the experiments. While the data and the software will be made available online, a complete description of the conducted experiments can be found in the supplementary material. 


\vspace{\cheating}
\subsection{Multimodal synchronization}
One critical issue when recording with several independent devices is synchronization, as  the same event is labeled by independent sensors with different timestamps. In the case of SALSA, we had to synchronize the eighteen badges worn by targets to the camera network. Assuming that there is no drift, we need to find the time-stamp mapping between each sociometric badge and the camera network. This problem reduces to determining the time-shift for each badge so that events are simultaneously observed by the badge and the camera network.

\begin{figure}[t]
 \centering
 \includegraphics[width = 0.9\linewidth]{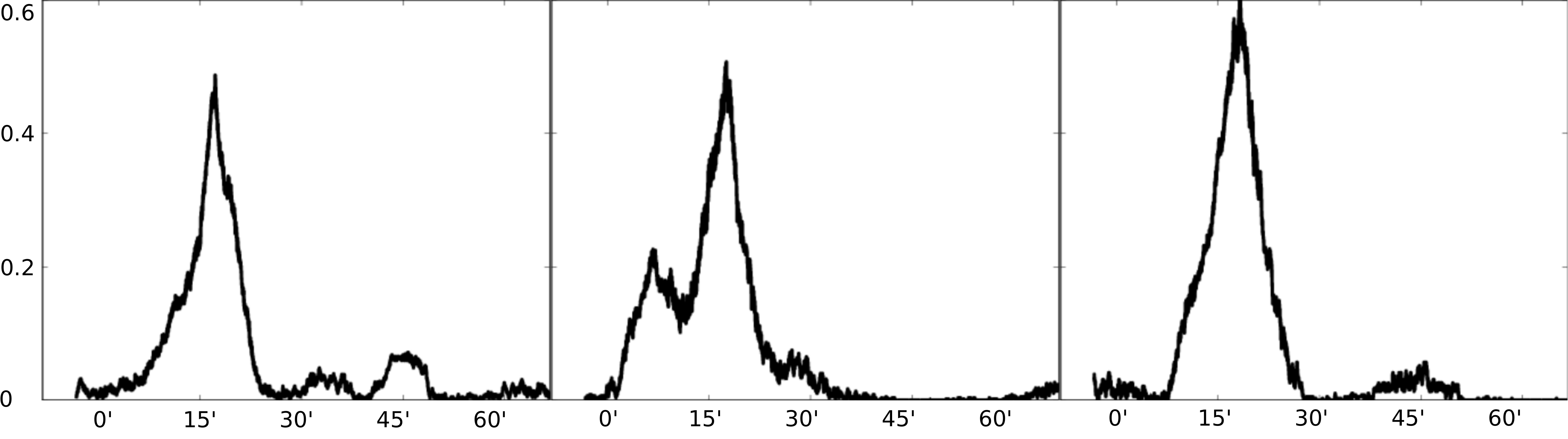}\vspace{-.05in}
 \caption{Synchronization procedure: Similarity scores for badge/target IDs $5$, $10$ and $16$ as a function of the time-shift-- a clear peak can be observed for all badges.}
\label{fig:sync}
\vspace{-0.4cm}
\end{figure}

Using the position and body pose annotations, we determined the set of potential infra-red 
detections, timestamped with respect to the cameras. By computing the similarity score between the potential and actual infra-red detections, we robustly estimated the temporal shift between each badge and the camera network. Computed scores for three of the badges, as a function of the shift are shown in Figure~\ref{fig:sync}. We can observe a clear peak in 
the badge's similarity score at the optimal time-shift. Computational details and obtained plots for all badges can be found in the supplementary material. 

\vspace{\cheating}
\subsection{Visual tracking of multiple targets}
Despite many advances in computer vision research, tracking individuals is still a unsolved problem.
In the particular case of SALSA, person tracking is challenging due to the presence of extreme and persistent occlusions. Some targets are difficult to distinguish from others using appearance features, and identity-preserving tracking required for multimodal behavior interpretation is further hindered by non-uniform scene illumination even when multiple views are available. State-of-the-art tracking-by-detection methods feature global appearance optimization~\cite{shitrit2014multi}, but require a sufficiently dense number of high-confidence detections across the whole sequence. However, target detection by itself is extremely challenging in such scenes even if leveraged through learning a set of detectors adapted to different occlusion levels~\cite{mathias2013handling}. We therefore considered a sequential Bayesian tracking approach without appearance model adaptation. The Hybrid Joint-Separable particle filter (HJS-PF) tracker~\cite{lanz2006approximate} was specifically developed for systematic occlusion handling at frame-level, and has been applied to tracking in social scenes~\cite{zen2010space,subramanian2013relationship,setti2013multi}.

\begin{figure*}
\centering
\includegraphics[width=0.9\textwidth]{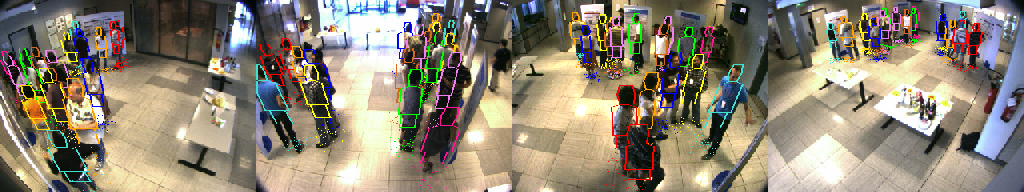}
\includegraphics[width=0.9\textwidth]{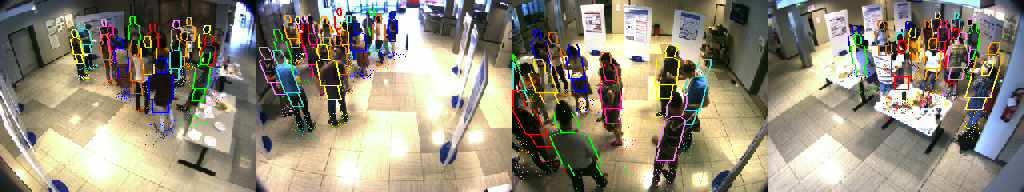}
\caption{Tracking on Part 1  - {\em Poster} (top row) and Part 2 - {\em Party} (bottom row). Best viewed in color.}
\label{fig:tracking}
\vspace*{-0.4cm}
\end{figure*}

HJS-PF represents targets' states in the scene based on ground locations, and applies a multi-target color likelihood with a first-order dynamical 
model to propagate a particle-set approximation of the posterior marginals for each target. In particular, the tracker exploits camera calibration 
information and a coarse 3D shape model for each target to explicitly model occlusion in the joint likelihood. While exact joint tracking  is 
intractable with increasing number of targets (exponential blow-up induced by the curse-of-dimension), it is shown in~\cite{lanz2006approximate} 
that 
single-target marginals can be updated under explicit occlusion reasoning with quadratic complexity, making the tracking of all 18 SALSA targets 
feasible. Furthermore, to prevent marginals from overlapping when targets have similar appearance-- a frequent failure mode leading to identity 
switches -- a Markov Random Field (MRF) defined over the targets' positions is added in the propagation. At each HJS-PF iteration, the propagation is 
solved via message-passing and the update combines HJS-PF likelihoods from each view independently. With a final resampling, the \textit{a 
posteriori} particle representation is obtained for each target. Details can be found in the supplementary material. 


We report HJS-PF tracking results on SALSA following the Visual Object Tracking Challenge (VOT 2013-14) evaluation protocol. The color model for each target was manually extracted from the initial 
part of the sequence where the target was free of occlusions, prior to tracking. These models were used for the whole sequence and were not re-initialized or adapted during tracking. HJS-PF was 
initialized for each target with the first available annotation, and tracking was performed at full frame rate (15 Hz) with 320$\times$240 resolution, while evaluation was done every 3 sec (or 
every 45 frames) consistent with the annotations. If the position estimate was over 70 cm from its reference, it was counted as a failure and the tracking of that target was re-initialized at the 
reference. Otherwise, the distance from the reference was accumulated to compute precision. In Table~\ref{tab:tracking}, the average precision (average distance from the references), per-target 
failure rate (\% of failures over 20K annotations), and frames-to-failure count (number of subsequent frames successfully tracked) are reported for 
the (i) first 30K frames (\textit{Poster}), (ii) the 
remaining 25K frames (\textit{Party}) and (iii) the total 60 minute recording. Our multi-thread implementation used in these experiments tracks the 18 targets using 50 particles per target at 7 fps on a 3 GHz PC. While overall precision is high considering space dimensionality, low image-resolution and high occlusion rate (cf. last row of table; to our best knowledge no comparable dataset exists 
for tracking evaluation), a sensible increase in failure rate is observed for the \textsl{Party} session. Indeed, in FCGs, persons tend to occupy every available space 
and exhibit a relaxed body posture such that they are hardly visible in some of the camera views. Also, targets more often bend their bodies to grab food and beverages, and illumination varies 
considerably over the scene impeding color-based tracking. However, low failure rate in the \textit{Poster} session where targets arrange themselves 
in a more orderly manner around posters indicates that occlusion handling is effective with the HJS-PF filter. A snapshot of the tracking results 
during the \textit{Poster} and the \textit{Party} scenarios is found in Figure~\ref{fig:tracking}. Based on these results, we identify some key 
elements requisite for FCG tracking: (i) perform ground tracking with explicit occlusion handling at frame level, (ii) extract discriminative signatures for 
each target to resolve identity switches (as in re-identification research), and (iii) learn the illumination pattern of the scene to adapt 
signatures locally to lighting conditions~\cite{mutlu2014constancy} (as in color constancy research). These may be cast into a global optimization framework~\cite{shitrit2014multi} and extended to multi-modal tracking.


 \begin{table}[t]
   \centering
   \caption{Mean tracking statistics and per-target occlusion rates for the four views.}\vspace{-.1in}
   \label{tab:tracking}
    \scalebox{1}{
   \begin{tabular}{lccc}
    \toprule
     & \textit{Poster} & \textit{Party} & All\\
    \midrule
     Precision (cm)  & 15.2 $\pm$ 0.1  & 20.1 $\pm$ 0.1 & 17.3 $\pm$ 0.1 \\
     Failure rate (\%) & 2.6 $\pm$ 0.1 & 9.6 $\pm$ 0.3 & 5.7 $\pm$ 0.2 \\
     Frames-to-failure & 1644 $\pm$ 63 &  439 $\pm$ 12 &  759 $\pm$ 21  \\
     \midrule
     {\em Occlusion (\%)} & {\em 28,35,22,26} & {\em 25,28,49,27} & {\em 27,32,34,27}\\
    \bottomrule
   \end{tabular}}
   \vspace{-0.4cm}
 \end{table}

 \vspace{\cheating}
\subsection{Head and body pose estimation from visual data} 
The estimation of the head and body pose is still an important research topic in the computer vision community. Specifically, when focusing on estimating the positions and head and body orientation of individuals in FCGs monitored by distant surveillance cameras, several challenges arise due to low resolution, clutter and occlusions. 
To demonstrate these challenges on SALSA, we considered the recent work 
of Chen \etal~\cite{chen2012we}, which is one of the few methods that jointly compute head and body orientation from low resolution images. In a nutshell, this algorithm consists of two phases. First, Histograms of Oriented Gradients (HoG) 
are computed from head and body bounding boxes obtained from training data. Then, a convex optimization problem that jointly learns two classifiers for head and body pose respectively is solved. Importantly, the classifiers are learned simultaneously, imposing consistency on the computed head and body classes so as to reflect human anatomic constraints (\textit{i.e.}, the body orientation naturally limits the range of possible 
head directions). The approach in~\cite{chen2012we} leverages information from both annotated and unsupervised data via a manifold term which imposes smoothness on the learned classification functions, typical of semi-supervised learning methods. In our experiments, only labeled data were used for training. 

The method proposed in~\cite{chen2012we} is monocular and considers $8$ classes (corresponding to an angular resolution of $45^o$) for both head and body classification. Therefore, to test it on SALSA, we also considered each camera view separately. In this series of experiments, the target head and body bounding boxes were obtained by manual annotation, and a subset of about 7.5K samples was employed (bounding boxes were not
available for targets going out of the field of view, see supplementary material). To compute visual features for both head and body, we used the 
UoCTTI variant of HoG descriptors~\cite{vedaldi2010vlfeat}. In our tests, \addnote[1-5-10]{1}{a small percentage of the frames ($1\%$, $5\%$, 
$10\%$)}~ were used for training, while the rest were used for testing. For performance evaluation, we used the mean angular error (in degrees) 
defined in~\cite{chen2012we} for computing head and body pose estimation accuracy. 


\begin{table}[t]
  \centering
  \caption{Head and body pose estimation error (degree).}\vspace{-.1in}
  \label{tab:headbodyvisual}
  \resizebox{0.99\linewidth}{!} {
  \begin{tabular}{cccccc}
  \toprule
$\%$ training data &   & view 1 &  view 2 & view 3 & view 4 \\
  \midrule
 \multirow{2}{*}{1$\%$} &  Head  & 45.7 $\pm$ 0.6 & 47.2 $\pm$ 0.3&  48.4 $\pm$ 0.8 & 49.5 $\pm$ 1.2 \\
  &  Body  & 49.3 $\pm$ 0.5  & 51.6 $\pm$ 0.9 & 51.2 $\pm$ 0.4 &  54.6 $\pm$ 0.8 \\
  \midrule
 \multirow{2}{*}{5$\%$}    &  Head  & 43.6 $\pm$ 0.5 & 46.2 $\pm$ 0.3&  46.4 $\pm$ 0.8 & 47.5 $\pm$ 0.9 \\
  &  Body  & 47.3 $\pm$ 0.5 & 49.4 $\pm$ 0.5 & 49.9 $\pm$ 0.5 &  52.5 $\pm$ 0.7 \\
  \midrule
 \multirow{2}{*}{10$\%$}    &  Head  & 42.2 $\pm$ 0.4 & 45.3 $\pm$ 0.3&  43.4 $\pm$ 0.8 & 44.9 $\pm$ 1.5 \\
  &  Body  & 45.4 $\pm$ 0.5 & 47.5 $\pm$ 0.8 & 48.7 $\pm$ 0.7 &  51.7 $\pm$ 0.5 \\
  \bottomrule
  \end{tabular}
  }
  \vspace{-0.4cm}
\end{table}

Experiments were repeated ten times with random training sets, and corresponding average error and standard deviation are reported in Table \ref{tab:headbodyvisual}. Despite many occlusions and the presence of clutter, a state-of-the-art pose 
classification approach achieves satisfactory performance (maximum error of around one class width). However, it is worth noticing that our experiments were performed with homogeneous training and test data, in contrast with the 
heterogeneous data employed in~\cite{chen2012we}. We expect a significant decrease in performance when heterogeneous training data are used for pose estimation. Also, errors observed for head pose are 
considerably smaller than for body pose over all four camera views-- this is because body pose classifiers are impeded by severe occlusions in crowded scenes. Precisely for this reason, previous 
works on F-formation detection from FCGs~\cite{cristani2011,setti2013multi,vascon} have primarily employed head orientation, even though body pose has been widely acknowledged as the more reliable cue 
for determining interacting persons. We believe that devising a multimodal approach also employing IR and bluetooth-based sensors for body pose estimation would be advantageous
as compared to a purely visual analysis, which was one of the primary motives for compiling the SALSA dataset.

\vspace{\cheating}
\subsection{Speaker recognition}
\label{sec:exp:speaker}
\begin{figure}[t]
	\centering
	\includegraphics[width=0.9\linewidth]{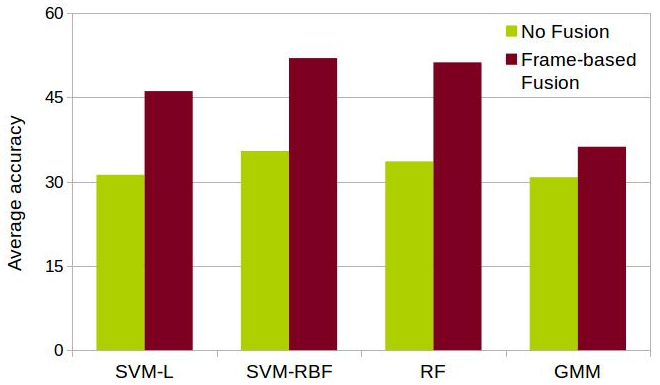}\vspace{-.1in}
	\caption{\label{fig:speaker_recognition}Mean speaker recognition accuracy with different methods on SALSA.}
\vspace{-0.4cm}
\end{figure}
Speaker recognition is a critical and fundamental task in behavior analysis from FCGs. Processing the auditory data in SALSA is challenging for several reasons. First, the recordings were carried out 
in a regular indoor space prone to reverberations and ambient noise. Second, 18 persons participated in the event and freely interacted with one another and therefore, the audio recordings 
consist of mixtures of speech signals emanating from different speakers. Third, the sociometric badges only retained part of the time-frequency representation, and thus high-performance speaker recognition 
is very challenging on this data. Finally, as relative positions of the speakers were constantly evolving, the speaker-to-microphone filter is 
not only unknown but highly time-varying, and thus very hard to estimate in practice. Indeed, current methods for speaker localization~\cite{alameda2014,canclini2013} or source 
separation~\cite{badeau2013,ozerov2010} are not designed for such a complex scenario.

Classical speaker recognition approaches build on Mel Frequency Cepstral Coefficients (MFCCs). Computed from the short-time frequency transform (STFT), MFCC have been shown to achieve a good balance between descriptive power, complexity and dimensionality~\cite{rabiner1993}. Four classifiers, namely, support vector machines with linear (SVM-L) and radial-basis function kernel (SVM-RBF), Gaussian mixture models (GMM) and random forests 
(RF) were employed for MFCC-based speaker recognition. In addition to the straightforward strategy of feeding the badge-specific MFCCs to classifiers, 
we also concatenated the MFCCs extracted from all badges at every frame to deal with time varying speaker-microphone relative locations. We refer to 
these two strategies as ``No Fusion'' (NF) and ``Frame-based 
Fusion'' (FBF) respectively. To create ground-truth labels, we annotated by visual inspection the ID of the speaker within a group of five persons interacting over a 15-minute duration during the poster session. We chose this group as the camera perspective and resolution allowed for reliable vision-based annotation.

Mean speaker recognition accuracies obtained upon five-fold cross validation are presented in Figure~\ref{fig:speaker_recognition}. We observe that the FBF strategy systematically outperforms the NF strategy. Focusing on FBF, we observe that RBF-SVM and random forests perform similarly, while doing much better than GMM and outperforming linear SVM. Also, as GMMs are known to be less effective in higher dimensional spaces, we computed principal components so as to $90\%$ variance in the FBF setting.


In light of these results, we outline directions for future work. First, due to the crowded nature of scenes involving FCGs, auditory analysis is highly challenging. Due to ambient noise, reverberations and multiple sound sources, recognizing the speaker from the badge audio 
data is challenging \textit{per se}. Nevertheless, performance increase observed with the FBF strategy suggests that a multi-modal approach can be effective, where tracking and pose estimates can facilitate multi-microphone based speech analysis. Finally, examining the badge data closely, most non-zero STFT coefficients are in the first nine frequency bins ($<300$Hz). Therefore, algorithms attempting speaker recognition on SALSA should design features to exploit this frequency range.

\begin{table}[t]
  \centering
  \caption{F-formation detection with ground-truth data.}
  \label{tab:ff-gt} \vspace{-.1in}
  \begin{tabular}{ccccccc}
   \toprule
	  & \multicolumn{3}{c}{Head} & \multicolumn{3}{c}{Body} \\
	  & Prec. & Rec. & F1 & Prec. & Rec. & F1 \\
  \midrule
      HVFF lin~\cite{cristani2011} & 0.56 & 0.72 & 0.63 & 0.59& 0.74 & 0.67 \\
      HVFF ent~\cite{SettiHC13} & 0.63 & 0.77 & 0.69 & 0.66 & 0.8 & 0.73 \\
      HVFF ms~\cite{setti2013multi} & 0.58 & 0.73 & 0.64 & 0.61 & 0.76 & 0.68 \\
      GC~\cite{setti2014} & 0.80 & 0.85 & 0.82 &0.82 & 0.85& 0.83 \\
   \bottomrule
  \end{tabular}
  \vspace{-0.4cm}
\end{table}

\vspace{\cheating}
\subsection{F-formation detection}
Detecting F-formations by visual observing crowded scenes is a challenging task. Several factors such as low video resolution, occlusions and complexities of human interactions hinder robust and accurate F-formation detection. We first considered four state-of-the-art vision-based approaches for individuating FCGs in SALSA. Specifically, we adopted (i) Hough voting~\cite{cristani2011} (HVFF-lin), (ii) its non-linear variant~\cite{SettiHC13} and (iii) multi-scale extensions~\cite{setti2013multi} (denoted as HVFF-ent and HVFF-ms) and (iv) the graph cut approach~\cite{setti2014} as associated codes are publicly available\footnote{http://profs.sci.univr.it/$\sim$cristanm/ssp/}. 

These approaches take the 
targets' positions and head pose as input, and compute F-formations independently for each frame. In particular, the Hough-voting methods work by generating a set of virtual samples around each target. These samples are candidate locations for the O-space center. By quantizing the space of all possible locations, aggregating samples in the same cell and finding the local maxima in the discrete accumulation space, the O-space centers and F-formations therefrom are identified. Oppositely, in the graph-cut algorithm, an optimization problem is solved to compute the O-space center coordinates.

We first evaluated the above F-formation detection approaches using ground-truth position and head and body pose annotations, and considered all the annotated frames. F-formation estimation accuracy is evaluated using precision, recall and
F1-score as in~\cite{cristani2011}.  In each frame, we consider a group as correctly estimated if at least $T\cdot|G|$ of the members are correctly found, and if no more than $1-T\cdot|G|$ non-members are wrongly identified, where $|G|$ is the cardinality of the F-formation $G$ and  $T=2/3$. Results are reported in Table \ref{tab:ff-gt}. Even the most accurate approach, \textit{i.e.} the graph-cut method, only achieves a F1-score of about $0.83$, clearly demonstrating the need of devising more sophisticated algorithms for detecting F-formations in challenging datasets such as SALSA. Moreover, it is worth noting that our results are consistent with the observations in previous works such as~\cite{cristani2011}, \textit{i.e.}, using the body pose is more advantageous than using head orientation for detecting a group of interacting persons.

In a second series of experiments, we evaluated the graph-cut approach using \textit{automatically} estimated head and body orientations from the multi-sensor badge data. Specifically, we considered annotations for the target positions, and estimated head and body pose from visual data with the method proposed in~\cite{chen2012we}. In these experiments, \addnote[multiview]{1}{HoG features extracted from head and body crops for the four camera views were concatenated and provided as input to the classifiers}. In this series of experiments, we only considered a subset of frames where \textit{all} the targets were in the camera field of view. To train the coupled head-body pose classifier, we used $1\%$ of the available samples as training data. Experiments were repeated ten times and the average performance is reported. We also integrated information from IR and audio sensors. Audio and IR are sparse observations, whilst visual data are available at every time-stamp. The likelihood of target $n$ addressing $m$ was estimated from audio and IR data. The maximum 
likelihood points to the person with whom $m$ (the addressee) is more likely to interact. The audio and IR observations correspond to the direction of 
the addresser ($n$). For integrating multiple angles, we simply considered their weighted average. Weights were tuned so as to maximize algorithm performance on a small validation set.

\begin{figure}[t]
\centering
   \includegraphics[width=0.9\linewidth]{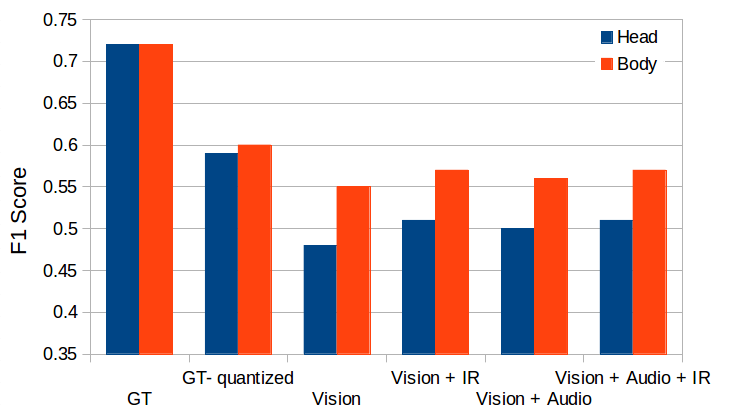}\vspace{-.1in}
   \caption{F-formation detection results (F1 score). Head and body pose were automatically estimated from visual, infra-red and audio data.}
   \vspace{-0.4cm}
\label{fig:ff-visual}
\end{figure}

The results of our experiments are reported in Fig.~\ref{fig:ff-visual}. Clearly, when the head and body pose are automatically computed from visual analysis, the performance significantly decreases with respect to the use 
of ground-truth (GT) annotations\footnote{Note that the accuracies with GT data 
reported in Fig.~\ref{fig:ff-visual} are different from those presented in Table~\ref{tab:ff-gt} since only a subset of frames is used in these experiments. In these considered frames, the scene is crowded since all 18 targets are inside the field of view.}.  Furthermore, by combining information from multiple modalities, we obtain a modest improvement with respect to using only visual data. Specifically, while jointly employing visual and infra-red data is advantageous with respect to exclusively employing the visual data, the integration of audio sensors provides minimum benefit. Finally, it is worth noting that a decrease in performance with respect to the ground-truth is also due to the angle quantization process. This can be observed by comparing the four leftmost bars in Fig.~\ref{fig:ff-visual}. Therefore, casting head and body pose estimation as 
a classification task (as typical of related works, where four/eight pose classifiers are used) appears to be insufficient for robustly detecting F-formations. Instead of considering classifiers, our experiments suggest that a better strategy entails casting head and body pose estimation as a regression task. 

\vspace{\cheating}
\subsection{Interaction networks and personality traits}
The SALSA behavioral data also allow for investigating the relationship between social dynamics and higher-level behavioral determinants such as personality traits. In this section, different from traditional works that have correlated audio-visual behavioral cues with the big-five traits, we show how the sociometric badge data can also be utilized for the same. To this end, we 
built three networks based on i) infra-red (IR) hits; ii) audio correlations\footnote{We computed the 
correlation between the badge STFT coefficients (normalized by the energy of the recording badge).}; and iii) group compositions from video-based F-formation annotations (GT).
To account for the dynamics, we selected windows of 60 and 120 seconds, and proceeded to build a multimodal graph for each window. For infra-red, the graph representing the 
$n$-th window has an edge between the target pairs whose badges detected an infra-red hit during the time period defined by $n$. For audio, we employed the correlations 
and added an edge between two targets if the corresponding correlation value is above a threshold (empirically set to 0.95). From the video data, we added an
edge between two targets when they were detected as being part of the same group.
Furthermore, we also built a static multimodal graph which encoded the entire sequence (equivalent to setting the window's duration equal to the sequence's duration).
From these networks, we extracted three basic classes of structural characteristics, i.e., \textit{centrality}, \textit{efficiency}, and \textit{transitivity}, and investigated how these 
characteristics are related to personality traits.

Inspired by previous studies \cite{kanfer1993, klein2004, wehrli2008, gloor2010}, we extracted the three standard measures of centrality proposed by Freeman: \textit{degree}, 
\textit{betweenness}, and \textit{closeness} centrality \cite{freeman1979}. These centrality measures can be divided into two classes: those based on the idea that the centrality of a node in a network 
is related to how close the node is to the other nodes (\eg, \textit{degree} and \textit{closeness} centrality), and those based on the idea that central nodes stand between others playing the role of 
intermediary (\eg, \textit{betweenness} centrality). Furthermore, we computed the network constraint~\cite{burt1992} for each individual; this measure provides an indication on how much the target's connections are connected with one another.

We also computed \textit{nodal} and \textit{local} efficiency for each node in the networks. The concept of \textit{efficiency}~\cite{latora2003} can be used to 
characterize how close to a `small world' a given ego-network is. Small world networks are a particular kind of networks that are highly clustered, like regular lattices, and have short 
characteristic path lengths like random graphs~\cite{watts1998}. The use of efficiency is justified by the hypothesis~\cite{lu2009} that the rate at which information flows within the network is influenced to some degree by the personality of the ego.

Finally, we extracted the \textit{transitivity} measure, which provides an indication of the clustering properties of the graph under analysis. Based on triads, \ie, triples of nodes in which either two (\textit{open}) or three (\textit{closed}) nodes are connected by an edge, \textit{transitivity} is defined as the ratio of the number of closed triads to the number of graph triads. In~\cite{wehrli2008}, Extraversion was found to negatively correlate with \textit{local transitivity}, while McCarty and Green~\cite{mccarty2005personality} found that agreeable and conscientious persons tend to have well-connected networks.

\begin{table}[t]
 \centering
  \caption{\label{tab:personality} Significant Pearson correlations between the big-five traits and IR and GT network features (* denotes $p<.01$).} \vspace{-.15in}
 \begin{tabular}{lll}
   \toprule
   \textbf{Trait} & \textbf{Feature} & \textbf{$R$} \\
   \midrule
   \multirow{2}{*}{Extr.} & Betw. Median Dyn-120-IR & .53 \\
    & Degree Std Dyn-60-IR & .53 \\
    \midrule
   \multirow{2}{*}{Agre.} & Degree Median Dyn-120-IR & .48 \\
    & Nodal Eff. Median Dyn-120-IR & .49 \\
    \midrule
   \multirow{4}{*}{Cons.} & Local Eff. Stat-IR & .52 \\
   & Nodal Eff. Median Dyn-60-GT & .49 \\
   & Nodal Eff. Median Dyn-120-GT & .56 \\
   & Trans. Mean Dyn-60-IR & .54 \\
   \midrule
   Em. St. & Trans. Median Dyn-60-IR & -.53 \\
   \midrule
   \multirow{6}{*}{Crea.} & Betw. Stat-GT & .49 \\
   & Clos. Mean Dyn-60-IR & -.49 \\
   & Degree Mean Dyn-60-GT & -.47 \\
   & Degree Mean Dyn-120-GT & -.50 \\
   & Loc. Eff. Stat-IR & -.51 \\
   & Nod. Eff. Stat-GT & -.49 \\
   & Trans. Stat-GT & -.73* \\
   \bottomrule
 \end{tabular}
\vspace{-0.45cm}
\end{table}


We conducted a preliminary statistical analysis (summarized in Table~\ref{tab:personality}) on the features derived from the interaction graphs 
described above, and investigated their associations with the personality data provided by SALSA. We only report associations that are significant at $p<.05$, 
unless otherwise stated. Unfortunately, we did not find statistical significant correlations between the personality traits and the 
auditory features. This issue will be subject of further investigation.

Extraversion, a personality trait lying in the tendency to behave in a way to engage and attract other people, and hence usually activated in 
situations such as social gatherings, was found to be significantly associated ($R=0.53$) with the standard deviation of the degree centrality 
computed on the 60 sec dynamic infra-red network. In other words, more extraverted targets appear to establish face-to-face interactions of variable 
duration, and thus engage with groups of diverse cardinality within the 1-minute windows under analysis. The expansion of the time window to 120 seconds 
provided additional insights: the higher a target scored on the Extraversion trait, the higher the median of his/her \textit{betweenness} centrality 
($R=0.54$). This suggests that the extravert targets in SALSA tended to act as \textit{brokers}, \ie, they served as connectors between clusters of 
people who had fewer face-to-face interactions.

Also, more agreeable subjects in SALSA were found to have a tendency towards engaging in face-to-face interactions with a higher number of people 
within highly connected clusters. Agreeableness was found to be significantly associated with the median degree centrality ($R=0.48$) and the median 
nodal efficiency ($R=0.49$) as computed on the 120 sec graphs. Emotional Stability was found to be negatively associated with median transitivity 
($R=-0.53$) on the 60 sec infra-red graphs-- this indicated that neurotic persons in SALSA tended to engage face-to-face during unbalanced interaction 
events.
  
Regarding Conscientiousness, several significant associations were found. In particular, from the 60 sec dynamic interaction networks built on infra-red 
data, we noted that conscientious subjects tended to participate in densely connected groups, as indicated by the positive association with mean 
transitivity ($R=0.54$). Thus, within the groups that naturally formed in the SALSA context, conscientious subjects took part mainly in those groups 
where the participants engaged more with each other. This fact is further confirmed by the significant association found on the static graph built on 
infra-red data between this trait and local efficiency ($R=0.52$). Interestingly, the median nodal efficiency extracted from the dynamic graphs built 
upon group annotations consistently shows similar associations with this trait ($R=0.56$ using a 120sec window, $R=0.49$ using 60 sec).

The social psychology literature does not offer many quantitative studies regarding the Creativity (or Openness to Experience) trait. Given this 
lack of information, the associations detected were remarkable, and deserve deeper investigation. This trait seemed to be associated with: i) local 
efficiency ($R=-0.51$) computed on the static face-to-face interaction network; ii) mean closeness centrality ($R=-.49$) computed on the 60 sec graphs 
built on infra-red; iii) betweenness centrality ($R=0.49$), nodal efficiency ($R=-0.49$), and transitivity ($R=-0.73, p<.001$) computed on the static 
group interaction network; and iv) mean degree in both 60 sec ($R=-0.47$) and 120 sec ($R=-.5$) graphs built from group annotations. Hence, creative 
persons seemed to participate overall in smaller and less connected networks than conservative targets.

While the above analyses present correlations between some of the behavioral cues that can be extracted from the SALSA data and high-level 
personality traits, we believe the entire gamut of information available can enable the study of both individual and group-level traits (\eg, 
\textit{dominance}). Also, unlike round-table meetings, which typically have an agenda based on which participants assume certain roles that may not 
relate to their actual personality, SALSA captures hedonistic and free-wheeling social interactions, which even if challenging to analyze, can provide 
a wealth of information about one's native behavior and personality.

\vspace{\cheating}
\section{Conclusions and future work} \label{Con}
Via extensive experiments, we have demonstrated how SALSA represents a rich but challenging dataset for analysis of FCGs. Vision-based analysis for target 
tracking, head and body pose estimation and F-formation detection evidenced the shortcomings of state-of-the-art methodologies when posed with cluttered scenes with persisting and extreme occlusions. 
However, additional sensors available as part of the sociometric badge were found to be helpful in cases where visual analysis was difficult-- in particular,  (i) the infra-red sensor which indicates both the proximity and body pose of the interacting counterpart was found to improve F-formation detection, (ii) both IR and visual cues were found to have significant correlations with the big-five personality traits, and (iii) improved speaker recognition with multi-badge speech data indicates the promise of additionally utilizing visual and accelerometer data to this end.  

Future research directions include: (a) developing new methodologies for robust audio processing in cluttered environments with many dynamic targets, (b) utilizing the bluetooth and accelerometer data for F-formation detection and personality trait recognition, and (c) designing tracking and head/body pose estimation algorithms capable of exploiting multimodal data. Given the extensive raw data and accompanying annotations available for analysis and benchmarking, we believe SALSA can spur systematic and intensive research to address the highlighted problems in a multimodal fashion in the near future. Evidently, SALSA would serve as a precious resource for the computer vision, audio processing, social robotics, social signal processing and affective computing communities among others.

\vspace{-.3cm}
{\footnotesize
\bibliographystyle{IEEEtran}
\bibliography{SALSAabrv,bibliograph}}
$ $\vspace{-2.5cm}
\begin{IEEEbiography}[{\includegraphics[width=1in,height=1.25in,clip,keepaspectratio]{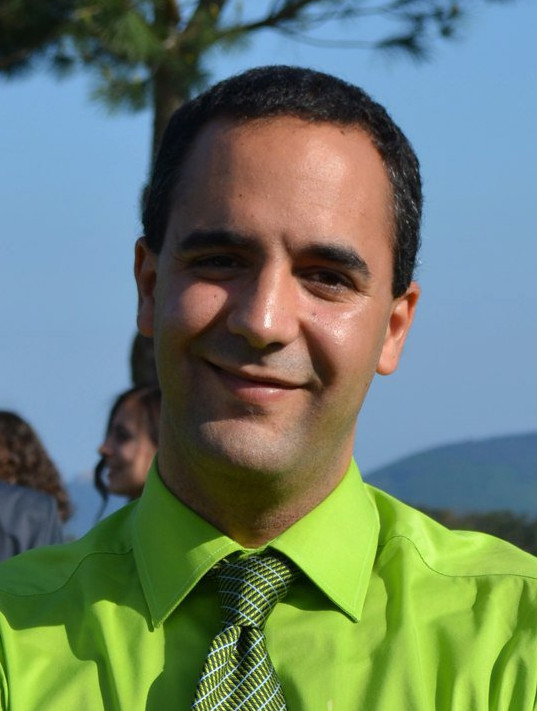}}]{Xavier Alameda-Pineda} received M.Sc. in mathematics, in 
telecommunications and in computer science. He worked towards his Ph.D. in mathematics and computer science in the Perception Team at INRIA, and obtained it from Universit\'e Joseph Fourier in 2013. 
He is currently a Post-Doctoral fellow at the University of Trento. His research interests are multimodal machine learning and signal processing for scene analysis.
\end{IEEEbiography}
\vspace{-0.6in}%
\begin{IEEEbiography}[{\includegraphics[width=1in,height=1.25in,clip,keepaspectratio]{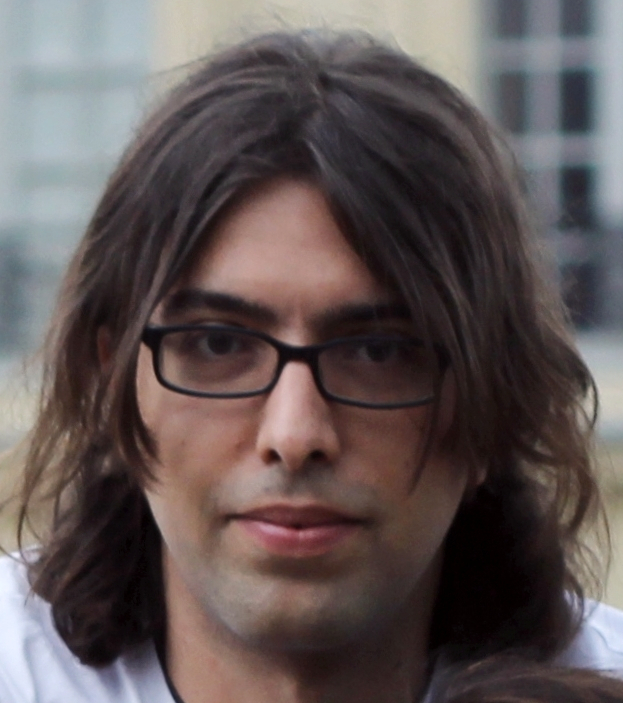}}]{Jacopo Staiano} is a researcher at LIP6, UPMC 
- Sorbonne Universit\'ees. He received his PhD from University of Trento in 2014. He was visiting scholar at the Ambient Intelligence Research Lab, 
Stanford University, at the Human Dynamics Lab, MIT Media Lab and at Telefonica I+D. His latest research efforts are on mobile computing and the 
economics of personal data. He received the Best Paper Award at ACM UBICOMP 2014.
\end{IEEEbiography}
\vspace{-0.6in}%
\begin{IEEEbiography}[{\includegraphics[width=1in,height=1.25in,clip,keepaspectratio]{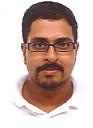}}]{Ramanathan Subramanian} received a PhD in Electrical and Computer engineering from the National 
University of Singapore in 2008. He is currently a Research Scientist at the Advanced Digital Sciences Center, Singapore. His research interests span human-centered computing, brain-machine 
interfaces, human behavior understanding, computer vision, and multimedia processing.
\end{IEEEbiography}
\vspace{-0.6in}%
\begin{IEEEbiography}[{\includegraphics[width=1in,height=1.25in,clip,keepaspectratio]{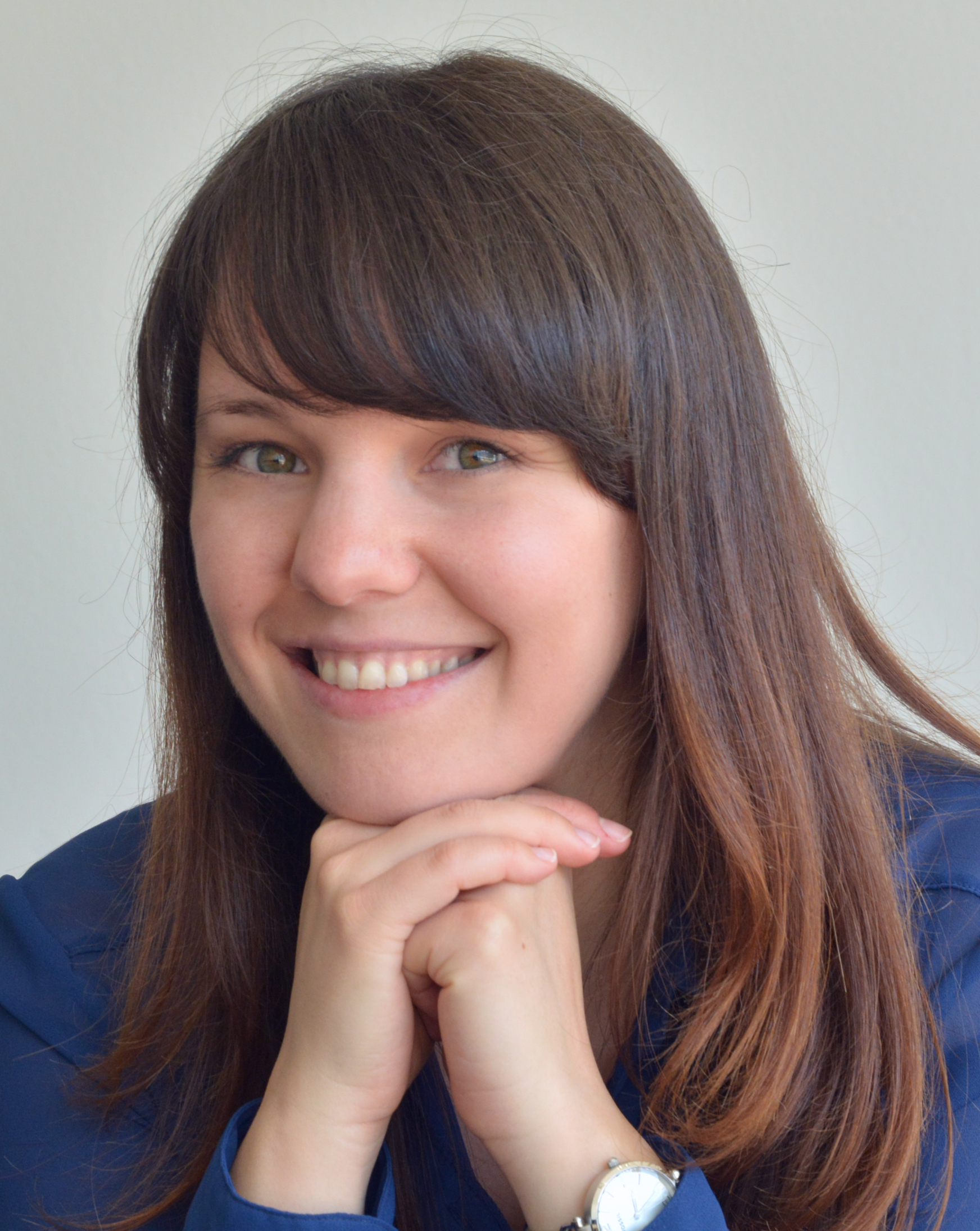}}]{Ligia Batrinca}  received the PhD degree in Cognitive and Brain Sciences in 2013 from the University of 
Trento, Italy. During her PhD, she was a visiting student at the Institute for Creative Technologies, USC. She is a Post-Doctoral fellow at the University of Trento. Her research interest are in the area of automatic human behavior analysis, including affective and social behaviors analysis 
and personality detection.
\end{IEEEbiography}
\vspace{-0.5in}%
\begin{IEEEbiography}[{\includegraphics[width=1in,height=1.25in,clip,keepaspectratio]{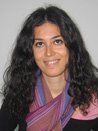}}]{Elisa Ricci} is an assistant professor at University of Perugia and a researcher at Fondazione 
Bruno Kessler. She received her PhD from the University of Perugia in 2008. She has since been a post-doctoral researcher at Idiap, Martigny and the Fondazione Bruno Kessler, Trento. She was also a 
visiting student at University of Bristol during her PhD. Her research interests are mainly in the areas of computer vision and machine learning.
\end{IEEEbiography}
\vspace{-0.6in}%
\begin{IEEEbiography}[{\includegraphics[width=1in,height=1.25in,clip,keepaspectratio]{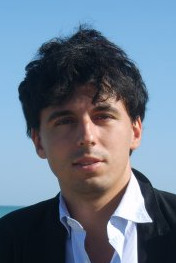}}]{Bruno Lepri} is a researcher at FBK and affiliate researcher at MIT 
Media Lab. At FBK, he is leading the Mobile and Social Computing Lab and he is vice-responsible of the Complex Data Analytics research line. In 2009, he received his Ph.D. in Computer Science 
from the University of Trento. His research interests include multimodal human behavior understanding, automatic personality recognition, network science, and computational social science.
\end{IEEEbiography}
\vspace{-0.6in}%
\begin{IEEEbiography}[{\includegraphics[width=1in,height=1.25in,clip,keepaspectratio]{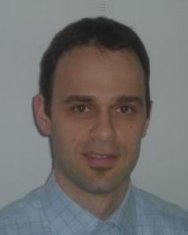}}]{Oswald Lanz} received a Masters in Mathematics and a Ph.D. in Computer Science 
from University of Trento in 2000 and 2005 respectively. He then joined Fondazione Bruno Kessler, and post his tenure in 2008, now heads the Technologies of Vision research unit of FBK. His research 
interests are mainly in the areas of computer vision and more recently in machine learning for vision. He holds two patents on video tracking.
\end{IEEEbiography}
\vspace{-0.5in}%
\begin{IEEEbiography}[{\includegraphics[width=1in,height=1.25in,clip,keepaspectratio]{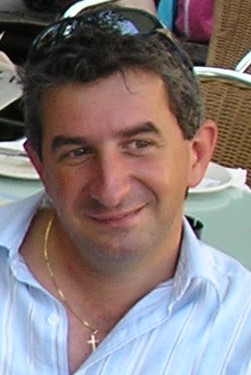}}]{Nicu Sebe} is a professor in the University of Trento, Italy, leading the research in  the areas of multimedia information retrieval and human behavior understanding. He was General Co-Chair of the IEEE FG Conference 2008 and ACM
Multimedia 2013, and Program Chair of the International Conference on Image and Video  Retrieval in 2007 and 2010 and ACM Multimedia 2007 and 2011. He is Program Chair of ECCV 2016 and ICCV 2017. He 
is a fellow of IAPR.
\end{IEEEbiography}

\end{document}